\documentclass{article}
\usepackage{spconf,amsmath,graphicx,hyperref}

\usepackage{newfloat}
\usepackage{listings}
\usepackage{amsmath}
\usepackage{booktabs}

\title{ENCORE: Entropy-Guided Cropping and Attention Regularization for Robust Vision--Language Understanding}
\name{Yuanhao Sun, Huawei Ji, Jiaxin Ding\textsuperscript{*}, Luoyi Fu, Xinbing Wang.\thanks{*Corresponding author. This work was supported by NSF China under Grant No. 62525209, 92579104, T2421002, T2542021.}}
\address{Shanghai Jiao Tong University, Shanghai, China\\}

\begin{document}
%\ninept
%

\maketitle

\begin{abstract}
Vision–Language Models (VLMs) perform well on diverse vision–language tasks, but transformer-based visual encoders split images into fixed-resolution sub-images, compromising object integrity in lightweight VLMs. Existing methods only focus on the visual modality and fail to dynamically preserve the integrity of prompt-relevant regions, limiting performance. In this work, we observe that the early‑layer image–text entropy of cross‑modal attention strongly correlates with answer grounding quality and task accuracy. Building on this finding, we propose \textbf{ENCORE}, an entropy-guided framework with two components: At inference, an \textbf{Entropy-based Cropping Strategy} (ECS) evaluates a small set of candidate crops and selects the one with minimal entropy, preserving contiguous regions relevant to the prompt. At training, \textbf{Entropy Regularization Training} (ERT) augments next‑token prediction with an entropy term that sharpens attention on key visual tokens while down‑weighting irrelevant ones. Experiments on ten VQA benchmarks show that ENCORE, fine-tuning only 0.14\% of parameters, achieves an average 1.43\% accuracy gain and state-of-the-art performance among recent 2B-parameter VLMs. Our code is released in https://github.com/baokou-fw2/ENCORE.

\end{abstract}

\begin{figure*}[t]
\centering
\includegraphics[width=\textwidth]{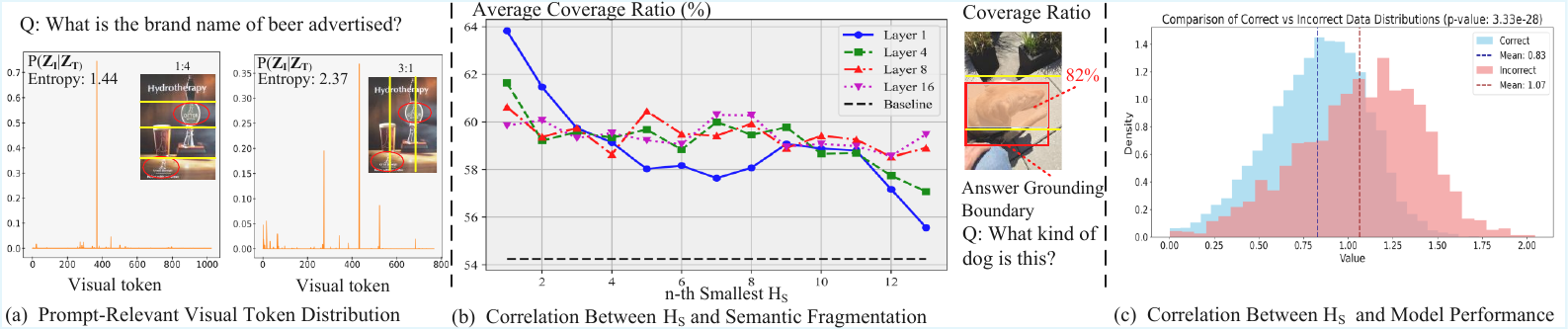}
\caption{
  \textbf{Insights into image–text entropy.} (a) Visualization of prompt-relevant visual-token probability distributions from VLM’s second layer; (b) Images cropped by ratio with the n-th smallest \(H_S\) and the maximum IoU(intersection over union) between patches and answer-grounding boundary (baseline: original aspect ratio) on VizWiz~\cite{chen2022grounding}. Results show that lower \(H_S\) indicates higher coverage ratio, corresponding to more accurate visual understanding; (c) Average \(H_S\) distributions over first four layers for correctly answered samples on TextVQA~\cite{singh2019towards}, HallB~\cite{guan2024hallusionbench}, SEED-2~\cite{li2024seed}, RWQA~\cite{grok15v} are lower than that of incorrectly answered samples, with a significantly statistic difference in T-test (\(t=-11, p=3.33 \times 10^{-28}<0.05\)).
}
\label{fig3}
\end{figure*}
\section{Introduction}
\label{sec:intro}
Recently, Vision–Language Models (VLMs)~\cite{wang2024emu3,zhang2024internlm,guo2025deepseek} have achieved remarkable performance across diverse vision–language tasks, demonstrating strong general reasoning ability. For fine-grained perception, visual encoders such as ViT~\cite{dosovitskiy2020vit} and SAM~\cite{kirillov2023segment} typically partition images into numerous high-resolution patches. However, this strategy often compromises the integrity of prompt-relevant regions—for instance, the word ‘Breakdown’ may be split into ‘Break’ and ‘down’, leading to semantic fragmentation~\cite{liu2024textmonkey,huang2024mini} and impairing holistic visual understanding.

Existing solutions fall into two categories: architecture optimization and image preprocessing. TextMonkey~\cite{liu2024textmonkey} employs sliding-window attention to maintain image continuity but still suffers from patch-based fragmentation~\cite{huang2024mini}. Preprocessing approaches~\cite{chen2024internvl,ye2023ureader,huang2024mini} leverage dynamic cropping, resizing, thumbnails, or multi-scale inputs, yet they remain limited to the visual modality and fail to ensure integrity of prompt-relevant regions, leaving efficient high-resolution inputs unresolved. To address this challenge, we examine how VLMs leverage textual prompts to interpret fragmented images. Since early layers are known to align visual and textual semantics effectively~\cite{zhang2025llava,zhang2024sparsevlm}, we compute cross-modal attention by applying a temperature-scaled softmax over CLIP similarities~\cite{radford2021learning}. Our analysis (Fig.~\ref{fig3}) reveals a clear trend: lower cross-modal entropy in early layers consistently corresponds to more complete answer-grounding coverage and higher VQA accuracy, highlighting entropy as a reliable indicator of semantic integrity.

Based on these findings, we propose ENCORE, an entropy-guided optimization framework for lightweight VLMs with two stages: \textbf{ECS} computes image–text entropy from early VLM layers over predefined aspect ratios, cropping by the smallest entropy to preserve contiguous prompt-relevant semantics; \textbf{ERT} uses this entropy as a self-supervised signal to help the model identify key prompt-relevant visual tokens during fine-tuning. ENCORE achieves SOTA among recent 2B-parameter VLMs. Our key contributions are as follows:

\begin{itemize}
  \item We introduce the image–text entropy—computed from early-layer cross-modal attention distributions—as a principled metric to quantify the degree of semantic fragmentation in prompt-relevant regions, and demonstrate its strong correlation with VLM performance.
\item We propose \textbf{ENCORE}, combining ECS for entropy-guided cropping to preserve prompt-relevant regions and ERT for entropy-regularized training to improve key visual token extraction.

  \item 
  With only 0.14\% of model parameters trainable via LoRA fine-tuning, 
  ENCORE achieves state-of-the-art performance across ten VQA benchmarks and exhibits strong generalization on recent 2B-parameter VLMs.
\end{itemize}

\section{Method}

We first derive the image–text entropy. Subsequently, we introduce the two primary components of the proposed ENCORE framework based on this discovery: Entropy‐Based Cropping Strategy (ECS) and Entropy Regularization Training (ERT). The overall architecture is illustrated in Fig.~\ref{model}.

\subsection{Image-Text Entropy}
In typical VLM pipelines, images and texts are first mapped into a shared embedding space for joint reasoning. However, partitioning images into high-resolution patches often splits contiguous object regions. Prior methods~\cite{liu2024textmonkey,huang2024mini} mitigate such semantic fragmentation through cropping or encoder designs, yet largely overlook the guiding role of textual prompts~\cite{zhang2025llava}. To leverage multi-modal information, we extract image tokens \(\mathbf{Z}_{I}\in\mathbf{R}^{N\times D}\) and text tokens \(\mathbf{Z}_{T}\in\mathbf{R}^{M\times D}\) from the \(k\)-th layer, and model the prompt-relevant attention distribution as:
\begin{equation}
    p(Z_{I}^{i}\mid \mathbf{Z}_{T}) = \frac{\exp\!\left(\tfrac{1}{M}\sum_{j=1}^{M} \mathbf{Z}_{I}^{i}\cdot(\mathbf{Z}_{T}^{j})^{T} / \tau\right)}{\sum_{k=1}^{N} \exp\!\left(\tfrac{1}{M}\sum_{j=1}^{M} \mathbf{Z}_{I}^{k}\cdot(\mathbf{Z}_{T}^{j})^{T} / \tau\right)},
\end{equation}

where \(M\) is the prompt length, \(N\) the number of visual tokens, and \(\tau\) the temperature. We then define the image–text entropy to measure fragmentation:
\begin{equation}
H_{S} = -\sum_{i=1}^{N} p(Z_{I}^{i}\mid \mathbf{Z}_{T}) \log\bigl(p(Z_{I}^{i}\mid \mathbf{Z}_{T}) + \varepsilon\bigr),
\label{ecs4}
\end{equation}
where \(\varepsilon\) is a small constant. This entropy serves as a unified indicator of the integrity of prompt-relevant visual regions, providing a quantitative basis for analyzing and mitigating semantic fragmentation in VLMs.

\subsection{ECS: Entropy-Based Cropping Strategy}

Based on the definition of image–text entropy \(H_{S}\) (Sec.~\ref{ecs4}) and its correlation with answer-grounding coverage (Fig.~\ref{fig3}), a smaller \(H_{S}\) corresponds to more contiguous preservation of prompt-relevant regions. Moreover, from a theoretical perspective, regardless of whether the prompt-relevant regions in an image are independent or distributed across multiple areas, the entropy can reflect their integrity within the range of candidate cropping ratios.
\begin{figure*}[t]
\centering
\includegraphics[width=\textwidth]{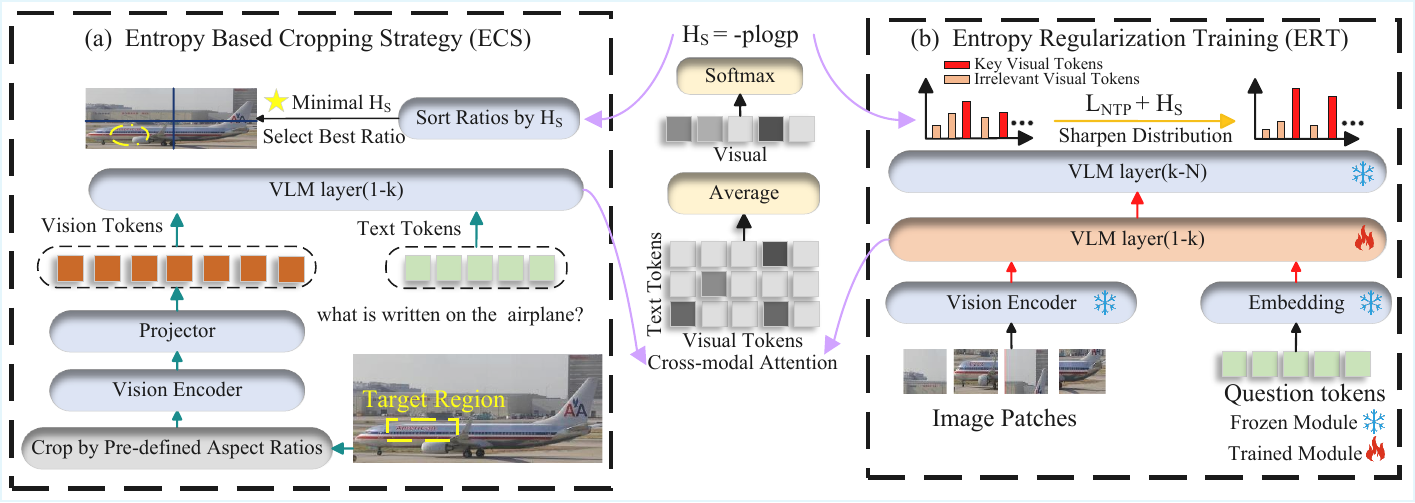}
\caption{
  \textbf{ENCORE framework.} (a) \textbf{ECS:} Compute early-layer cross-modal similarities between visual tokens and prompt, derive prompt-relevant distribution, compute its entropy, select minimal-entropy crop ratio to preserve contiguous prompt-relevant semantics;
  (b) \textbf{ERT:} During fine-tuning, augment next-token prediction loss with entropy regularizer to encourage sharpening cross-modal attention for accurate key visual token extraction from fragmented high-res inputs.
}
\label{model}
\end{figure*}

As illustrated in Fig.~\ref{model}, ECS computes \(H_{S}\) from the first \(k\) layers of the VLM, selects the cropping ratio with minimal entropy, and integrates the resulting crops with the originals to form the final visual input.

\vspace{-6mm}
\subsection{ERT: Entropy Regularization Training}
In VQA tasks, VLMs are typically trained with next-token prediction~\cite{chen2024far}, which lacks explicit suppression of mismatched visual–text pairs. When images are partitioned into many high-resolution patches, this often leads to difficulty in distinguishing key visual tokens from irrelevant ones. 
\begin{table*}[ht]
\centering
\setlength{\tabcolsep}{1mm} % 设置列间距为1mm
\caption{Comparison results with SOTA VLMs. ENCORE achieves the best results among 2B-parameter VLMs and surpass some larger VLMs in several benchmarks.}
\begin{tabular}{l|l|cccccccccc}
\toprule
Model & Size & SEED-2&Info & Text & OCRB & CCB & RWQA & HallB & POPE &\(\text{MMB}_{\text{v1.1}}^{\text{EN}}\)&MMStar\\
\midrule
Deepseek-VL~\cite{lu2024deepseek} & 2B  & 43.3& 21.0 & 57.7 & 414 & 37.6 & 49.7 & 27.6 & 85.9& 63.8 & 39.9\\
Qwen2-VL~\cite{wang2024qwen2} & 2B & 62.4& 65.5 & 79.4 & 809 & 53.5 & 62.6 & 41.7 & 87.8& 72.7& 47.8 \\
mPLUG-Owl3~\cite{ye2024mplug} & 2B  & 44.6 & 29.0& 59.2 & 434 & 33.7 & 55.5 & 25.1 & 85.8 & -& 41.2\\
InternVL-2~\cite{chen2024far} & 2B  & 60.0 & 58.0& 74.7 & 779 & 74.3 & 57.3 & 37.9 & 85.2& 70.2 & 50.2\\
Mini-monkey~\cite{huang2024mini}& 2B & 61.2& 60.0 & 75.7 & 802 & 75.5 & 57.5 & 38.7 & 86.7& 68.9& 50.4 \\
InternVL-3~\cite{chen2024expanding} & 2B & 64.6 & 66.1& 77.0 & 831 & 74.9 & 64.3 & 41.8 & 89.6 & 78.6&60.7\\
ENCORE(ours) & 2B &\textbf{65.5}&\textbf{68.0}&\textbf{79.8}&\textbf{845}&\textbf{77.9}&\textbf{66.3}&\textbf{43.1}&\textbf{90.2}&\textbf{79.2}&\textbf{61.1}\\
\bottomrule
\end{tabular}
\end{table*}

To address this, \(H_S\) from the k-th layer during the fine-tuning stage serves as an entropy regularization term. As shown in Fig.~\ref{model}, \(H_S\) encourages higher attention on key visual tokens and lower attention on irrelevant tokens. Specifically, we reconstruct the training loss as:
\begin{equation}
\mathcal{L} \;=\; 
- \sum_{\ell=1}^{L} \log p(y_\ell \mid y_{<\ell}, x_{\le \ell})
\;+\;\alpha\,H_S,
\label{eq:ert_loss}
\end{equation}
where $\alpha>0$ controls the regularization strength. For prompt-relevant tokens ($p_i>1/e$), minimizing \(H_S\) reinforces their probabilities in line with cross-entropy, while for irrelevant tokens ($p_i<1/e$), the self-information term suppresses noisy pairs, resembling negative sampling in contrastive learning. These effects together sharpen attention toward key visual tokens in fragmented inputs.

\section{Experiment}
\subsection{Implementation Details}
Following previous work~\cite{chen2024internvl}, we fine-tune InternVL3-2B on the original training datasets and test on ten general multimodal understanding benchmarks of InternVL3-2B including TextVQA
%~\cite{singh2019towards}, 
, OCRBench%~\cite{liu2023hidden}, 
, SEED2-Plus and 
InfoVQA%~\cite{mathew2022infographicvqa}, 
, CCBench%~\cite{liu2024mmbench}, 
, RealWorldQA%~\cite{grok15v},
, HallBench%~\cite{guan2024hallusionbench}, 
, POPE%~\cite{li2023evaluating}, 
, MMBench-v1.1-EN
 and 
MMStar%~\cite{chen2024we}. 
. The base learning rate is 4e-8. \(\tau\) is set as 0.1 in ERT and 1 in ECS. Layer k is set as 2 in ERT and 1 in ECS. \(\varepsilon\) is selected as 1e-32. \(\alpha\) is set as 0.1. Cropping range for ECS is selected from 1 to 4.
\begin{table}[ht]

\centering
\caption{Comparison of ECS and other cropping strategies. Results are tested with InternVL3-2B.}
\resizebox{\columnwidth}{!}{%
\begin{tabular}{l|cccc}
\toprule
Strategy & TextVQA & OCRBench & RWQA & CCBench  \\
\midrule
DHR~\cite{chen2024far} & 77.0 & 831 & 64.3 & 74.9 \\
FHR~\cite{li2024monkey}  & 76.9 & 825 & 63.6 & 73.5 \\
MSAC~\cite{huang2024mini} & 78.8 & 834 & 63.0 & 75.0 \\
RTC~\cite{lu2025internvl} & 77.2 & 827 & 63.9 & 74.8 \\
ECS (ours) & \textbf{79.1} & \textbf{837} & \textbf{65.1} & \textbf{75.3} \\
\bottomrule
\end{tabular}%
}

\label{ablation_crop}
\end{table}
\subsection{Comparison of Other VLMs}

The results show that on text-rich datasets such as TextVQA and OCRBench, ENCORE surpasses InternVL3-2B by 2.8\% and 1.4\%, confirming that ERT enhances capture of question-relevant text. Beyond OCR, ENCORE also improves RealWorldQA (+2.0\%) and CCBench (+2.4\%), showing that ECS effectively preserves prompt-relevant regions. Moreover, ENCORE consistently outperforms all 2B-parameter baselines, reaching new highs on SEED-2 (65.5), InfoVQA (68.0), HallBench (43.1), POPE (90.2) and MMBench(79.2). 

Overall, ENCORE delivers broad gains with only 0.14\% trainable parameters and minimal overhead, achieving an excellent trade-off between efficiency and accuracy.

\begin{figure}[t]
\centering
\includegraphics[width=\columnwidth]{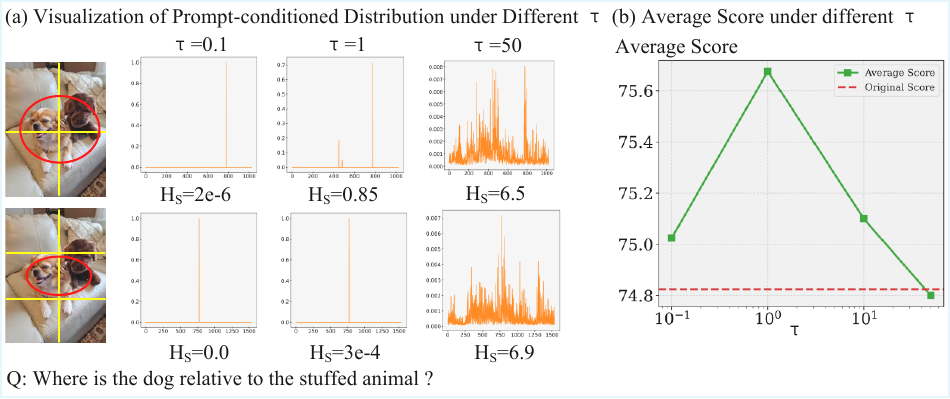} 
\caption{\textbf{Ablation study of $H_S$ under varying temperature $\tau$.} (a) Visualization of $\tau$'s impact. (b) shows the $\tau$'s impact on average score.}
\label{fig4}
\end{figure}

\vspace{-2mm}
\subsection{Ablation Study of ECS}
In this section, we evaluate the generalization of the Entropy‐Based Cropping Strategy (ECS) across different models and compare it with alternative cropping methods.

First, we analyze the VLM layer for computing \(H_S\): the first or second layer delivers comparable performance, while deeper layers show a marked drop, so we adopt the first layer for subsequent entropy calculations. Increasing \(\tau\) from 0.1 to 50 (Fig.~\ref{fig4}(a)) shifts the prompt-relevant token distribution from sharp to uniform, driving \(H_S\) upward; model performance peaks at \(\tau=1\) (Fig.~\ref{fig4}(b)). Moreover, with the max patch count set to 6 (as in InternVL3-2B~\cite{chen2024expanding}) and \(H_S\) computed via the first layer, average inference time only increases FLOPs by 7.2\% and inference latency by 10\%, demonstrating that entropy-guided cropping provides performance gains with relatively low burden.

Then, we compare ECS with several alternative cropping strategies (Table~\ref{ablation_crop}). 
On benchmarks such as OCRBench and TextVQA, MSAC performs comparably to ECS, likely because these datasets emphasize OCR tasks with limited prompt semantics. 
However, on benchmarks with complex visual semantics—such as SEED-2 and RWQA—MSAC and other baselines degrade performance, whereas ECS consistently improves accuracy. Although ECS introduces additional crops, the overall computational overhead remains modest.
\subsection{Ablation Study of ERT}
\begin{figure}[t]
  \centering
  \includegraphics[width=\columnwidth]{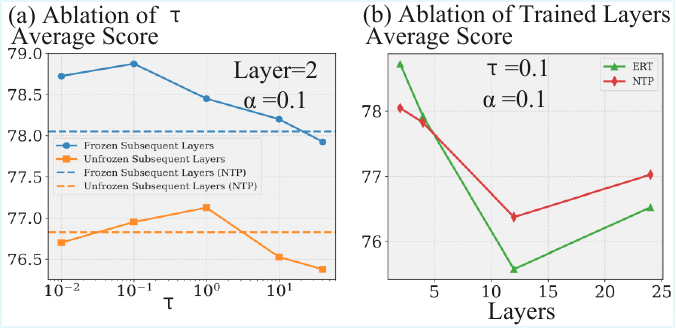}
  \caption{
    \textbf{Ablation study of Entropy Regularization Training (ERT).}
    (a) shows average score on POPE, OCRBench, RWQA and CCBench; In (a)–(b), NTP denotes next‑token‑prediction fine‑tuning.
  }

  \label{fig5}
\end{figure}

We select the POPE, RWQA, OCRBench, and CCBench benchmarks to systematically probe four factors influencing Entropy Regularization Training (ERT): whether to unfreeze subsequent layers, the choice of temperature coefficient \(\tau\), the number of fine-tuned layers, and the entropy regularization weight \(\alpha\). To isolate the contribution of ERT and its plug-and-play flexibility, ablation models are trained and tested with dynamic high-resolution cropping strategy according to original setting of InternVL3~\cite{chen2024far}.

Fig.~\ref{fig5}(a) shows that freezing all subsequent layers yields superior accuracy, suggesting that unfreezing deep layers disrupts the reasoning capabilities established by InternVL3’s Mixed Preference Optimization pretraining~\cite{wang2024enhancing}. Furthermore, a small temperature (\(\tau<1\)) sharpens the prompt-relevant visual‐token distribution, enhancing token discrimination and overall performance. In Fig.~\ref{fig5}(b), with the total number of trainable parameters held constant, ERT is most effective when applied to the initial layers, whereas fine‐tuning deeper layers underperforms the standard NTP.

\begin{figure}[t]
  \centering
  \includegraphics[width=\columnwidth]{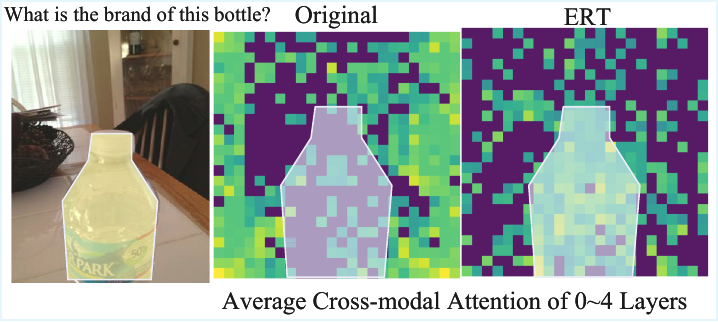}
  \caption{
    \textbf{Visualization of cross-attention matrices in layer 2 after Entropy Regularization Training (ERT).}
  }
  \label{fig6}
\end{figure}

Fig.~\ref{fig6} visualizes cross-attention matrices before and after ERT in the early layers: without ERT, attention is dispersed across regions, while ERT concentrates attention on key visual tokens relevant to the prompt, which aligns with ERT's role in improving key visual token extraction.
\vspace{-2mm}
\section{Conclusion}
To address semantic fragmentation in lightweight VLMs, we propose ENCORE—an entropy-guided framework that preserves prompt-relevant regions and sharpens key visual token extraction. Experiments on multiple VQA benchmarks show that ENCORE achieves state-of-the-art performance among 2B-parameter VLMs, demonstrating its effectiveness.

\newpage

% References should be produced using the bibtex program from suitable
% BiBTeX files (here: strings, refs, manuals). The IEEEbib.bst bibliography
% style file from IEEE produces unsorted bibliography list.
% -------------------------------------------------------------------------
\small  % 比默认小一号

\bibliographystyle{IEEEtranS}  % S 表示 simplified
\bibliography{strings,refs}

\end{document}